\documentclass{article} %
\usepackage{amsmath,amsfonts,bm}

\def\eqref#1{equation~\ref{#1}}
\def\1{\bm{1}}

\DeclareMathAlphabet{\mathsfit}{\encodingdefault}{\sfdefault}{m}{sl}
\SetMathAlphabet{\mathsfit}{bold}{\encodingdefault}{\sfdefault}{bx}{n}

\DeclareMathOperator*{\argmin}{arg\,min}

\usepackage{booktabs}
\usepackage{microtype}
\usepackage{graphicx}
\usepackage{subfigure}
\usepackage{hyperref}

\usepackage[accepted]{icml2023} 
\usepackage{amsmath,amsthm,mathtools,amsfonts,amssymb}

\usepackage{url}
\usepackage{algorithm}
\usepackage{algorithmic}
\usepackage{multicol}
\usepackage{multirow}
\usepackage{wrapfig}
\usepackage{xcolor}

\usepackage{ulem}

\theoremstyle{plain}
\newtheorem{theorem}{Theorem}[section]
\newtheorem{proposition}[theorem]{Proposition}

\theoremstyle{definition}

\theoremstyle{remark}

\icmltitlerunning{NeuralStagger: Accelerating Physics-constrained Neural PDE Solver with Spatial-temporal Decomposition}
\begin{document}

\twocolumn[
\icmltitle{NeuralStagger: Accelerating Physics-constrained Neural PDE Solver with Spatial-temporal Decomposition}

\icmlsetsymbol{equal}{*}

\begin{icmlauthorlist}
\icmlauthor{Xinquan Huang}{equal,uni}
\icmlauthor{Wenlei Shi}{equal,comp}
\icmlauthor{Qi Meng}{comp}
\icmlauthor{Yue Wang}{comp}
\icmlauthor{Xiaotian Gao}{comp}
\icmlauthor{Jia Zhang}{comp}
\icmlauthor{Tie-Yan Liu}{comp}
\end{icmlauthorlist}

\icmlaffiliation{uni}{King Abdullah University of Science and Technology, work done during an internship at Microsoft Research AI4Science}
\icmlaffiliation{comp}{Microsoft Research AI4Science}

\icmlcorrespondingauthor{Xinquan Huang}{xinquan.huang@kaust.edu.sa}
\icmlcorrespondingauthor{Wenlei Shi}{wenlei.shi@microsoft.com}
%\icmlkeywords{Machine Learning, ICML}

\vskip 0.3in
]
%\printAffiliationsAndNotice{}  % leave blank if no need to mention equal contribution
\printAffiliationsAndNotice{\icmlEqualContribution} % otherwise use the standard text.

\begin{abstract}
Neural networks have shown great potential in accelerating the solution of partial differential equations (PDEs). 
Recently, there has been a growing interest in introducing physics constraints into training neural PDE solvers to reduce the use of costly data and improve the generalization ability.
However, these physics constraints, based on certain finite dimensional approximations over the function space, must resolve the smallest scaled physics to ensure the accuracy and stability of the simulation, resulting in high computational costs from large input, output, and neural networks.
This paper proposes a general acceleration methodology called NeuralStagger by spatially and temporally decomposing the original learning tasks into several coarser-resolution subtasks. 
We define a coarse-resolution neural solver for each subtask, which requires fewer computational resources, and jointly train them with the vanilla physics-constrained loss by simply arranging their outputs to reconstruct the original solution.
Due to the perfect parallelism between them, the solution is achieved as fast as a coarse-resolution neural solver.
In addition, the trained solvers bring the flexibility of simulating with multiple levels of resolution. 
We demonstrate the successful application of NeuralStagger on 2D and 3D fluid dynamics simulations, which leads to an additional $10\sim100\times$ speed-up. Moreover, the experiment also shows that the learned model could be well used for optimal control. 
\end{abstract} 

\section{Introduction}
\label{introduction}
Partial differential equations (PDEs) are the critical parts of scientific research, describing vast categories of physical and chemical phenomena, e.g. sound, heat, diffusion, electrostatics, electrodynamics, thermodynamics, fluid dynamics, elasticity, and so on.
In the era of artificial intelligence, neural PDE solvers, in some works called neural operators, are widely studied as a promising technology to solve PDEs ~\cite{guo2016convolutional,zhu2018bayesian,hsieh2018learning,bhatnagar2019prediction,bar2019learning,berner2020numerically,li2020multipole,li_fourier_2020,Um2020solver,pfaff2020learning,lu2021learning,wang2021learning,kochkov2021machine}. 
Once the neural solver is trained, it can solve unseen PDEs with only an inference step, multiple magnitudes faster than that with traditional numerical solvers. 
Recently, several works have introduced physics constraints in training the neural PDE solvers in order to reduce the use of costly data and improve the generalization ability.
They define the physics-constrained loss with certain finite dimensional approximations to transform the PDEs into algebraic equations, which are further used to define the loss function \cite{zhu_physics-constrained_2019,geneva_modeling_2020,Wandel2020,shi_lordnet_2022}.
However, to ensure stability and accuracy, they must define the loss in a relatively high resolution to resolve the smallest-scale physics in the PDE, resulting in huge input and output as well as increased neural network size.
The solution by the neural network inference might still be slow, but it seems impossible to get further accelerations as the bottleneck comes from the input and output complexity. 

\begin{figure*}[t]
    \centering
    \includegraphics[width=0.75\textwidth]{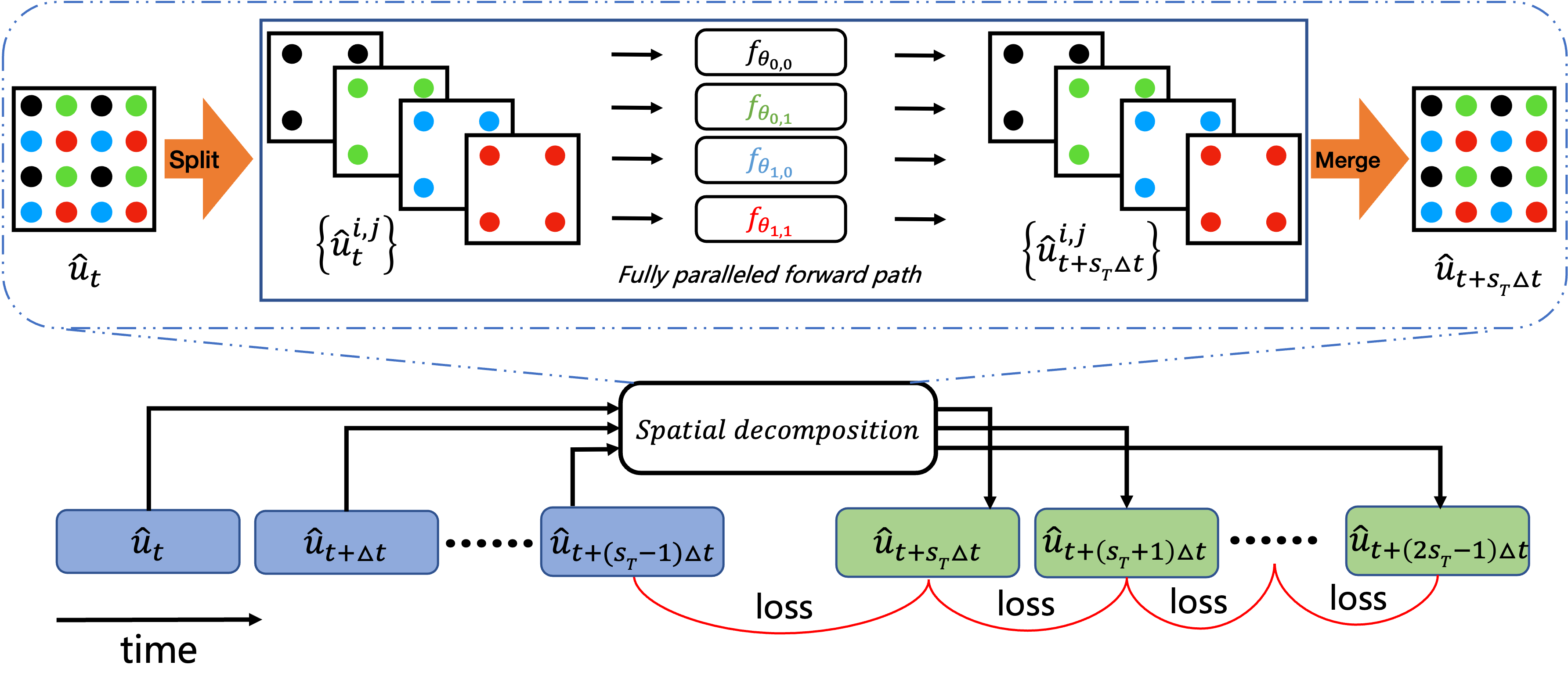}
    \vspace{-12pt}
    \caption{The training pipeline of NeuralStagger. \textbf{Top}: the spatial decomposition that splits the field into several pieces of coarse-resolution fields; \textbf{Bottom}: the temporal decomposition combined with spatial decomposition to construct the physics-constrained loss.}
    \label{fig:sp-strategy}
\end{figure*}
In this paper, we propose a simple methodology called NeuralStagger to jump out of the dilemma. 
The basic idea is to decompose the original physical fields into several coarser-resolution fields evenly. Then we jointly train a lightweight neural network to predict the solution in each coarse-resolution field respectively, which can be naturally a coarse-resolution neural solver to the original PDE. 
We design the decomposition rules so that the outputs of these lightweight networks can reconstruct the solutions in the original field with simple arrangements. 
For ease of reading, here and also in most parts of the paper, we illustrate the decomposition methodology in the 2-dimensional example with regular mesh and finite difference approximation.
Figure \ref{fig:sp-strategy} (top) shows the physical field in a $4\times 4$ mesh is decomposed into 4 coarser-resolution fields, each of which is handled by a small neural network.
We could also do similar things along the temporal dimension, as is shown in Figure~\ref{fig:sp-strategy} (bottom).
The group of coarse-resolution solvers, as well as the decomposition and reconstruction operations, can be seen as an end-to-end neural PDE solver, which can be trained with the physics-constrained loss that resolves small-scale physics in a sufficiently high resolution.
Because the neural networks can run in parallel, the original simulation is achieved as fast as a coarse-resolution neural solver.
In addition, the trained neural networks can predict the PDE's solution in various levels of resolution, ranging from the resolution of the individual coarse-resolution solver to the resolution of the physics-constrained loss by the combination of all these solvers. 
We believe that such flexibility is vital in balancing the computational resources and the resolution.

We demonstrate the effectiveness of the NeuralStagger in the Navier-Stokes equation with three parametric settings, e.g., periodic boundary conditions with varied initial conditions, lid-driven cavity boundary conditions with varied initial conditions, and the flow around the obstacle with varied obstacles and initial conditions. We find that with NeuralStagger, the learned networks can conduct accurate and stable simulations with $20\sim 400$ fold decrease on the computational load per GPU card or practically $10\sim100\times$ speed-up over SOTA neural PDE solvers. 
In addition, we demonstrate that they can accurately tackle the optimal control task with auto-differentiation.

Our contributions can be summarized in three parts:
\begin{itemize}
    \item We propose a general methodology called NeuralStagger to accelerate neural PDE solving by spatially and temporally decomposing the learning task and running a group of coarse-resolution solvers in parallelism.
    \item The learned network group can provide solutions in multiple resolutions from the coarsest one by a single network to the original resolution, which provides the flexibility to balance the computational resources and the resolution.
    \item We demonstrate that the methodology leads to $10\sim100\times$ speed-up over SOTA neural PDE solvers as well as the efficient solution on optimal control.
\end{itemize}

In the following sections, we first briefly summarize the related works in Section~\ref{related_work} and then introduce the preliminaries and the proposed NeuralStagger in Section~\ref{methodology}. To showcase the efficiency and accuracy of the proposed method, we present the settings of the experiments and results in Section~\ref{experiments}. Finally, we conclude and discuss the future work in Section~\ref{conclusion}.

\section{Related Work}
\label{related_work}
\textbf{Numerical methods.} The concept of stagger has been used in several classical methods, e.g., the Leapfrog integration scheme \cite{birdsall1985plasma} and the staggered grid method \cite{harlow1965numerical}. However, NeuralStagger is fundamentally different from these methods in both targets and technical details. 
NeuralStagger is proposed to accelerate neural PDE solving under certain physics-constrained loss, while the classical methods tell how the continuous PDE can be discretized and solved with algebra, which can be naturally used to define the physics-constrained loss.
As you would see in Section~\ref{sec:flow-around}, we leverage staggered grid method to define the physics-constrained loss in the flow around obstacles case.
In addition, the ways of decomposition are also different. The Leapfrog integration scheme updates positions and velocities at staggered time points, giving nice properties like time-reversibility and second-order accuracy; the staggered grid method stores the scalar variables in the cell centers and vector variables at the cell faces. In contrast, each coarse-resolution solver in NeuralStagger updates all the variables spatially in the same grid and temporally at the same time points.

\textbf{Neural PDE solvers.} 
The neural PDE solver learns to solve a parametric PDE with merely an inference step, which is much faster than the numerical methods.
Many impressive works have been done to improve the neural solver for parametric PDEs in terms of neural network design, e.g., convolutional neural network
~\cite{guo2016convolutional,tompson2017accelerating,bhatnagar2019prediction}, graph neural networks~\cite{pfaff2020learning}, the multipole graph kernel~\cite{li2020multipole}, Fourier neural operators~\cite{li_fourier_2020,guibas2021adaptive}, deepOnet~\cite{lu_deeponet_2021}, the message passing neural network~\cite{brandstetter2022message}, Clifford neural networks~\cite{brandstetter2022clifford} and so on.
After being trained with pre-generated simulated data and labels, they can solve the PDE several magnitudes faster than conventional numerical solvers with competitive accuracy. 
Recently there are raising concerns about the cost of collecting training data and the generalization ability, so several works have introduced the physics-constrained loss for training.  
For example, \citet{wang2021learning} combined the DeepOnet with a physics-informed way to improve the sample efficiency. 
\citet{zhu_physics-constrained_2019} proposed physics-constrained loss for high-dimensional surrogate modeling and \citet{geneva_modeling_2020} introduced the use of a physics-constrained framework to achieve the data-free training in the case of Burgers equations. 
\citet{Wandel2020, Wandel2021} proposed the physics-constrained loss based on the certain approximation of the Navier-Stokes equation to solve fluid-like flow problems. \citet{shi_lordnet_2022} proposed a general physics-constrained loss called mean square residual (MSR) loss as well as a neural network called LordNet for better performance. 
However, the physics-constrained loss by certain approximation requires the approximation to be sufficiently close to the continuous version, resulting in a relatively high-resolution discretization. 
Thus in complex and large-scale problems, the neural solver must be large enough for expressiveness and its inference would still be slow.
Although some works~\cite{wang2021learning} directly calculate the derivatives via back-propagation through the neural network, they are known to have similar training problems as PINN, e.g., converging to trivial solutions. 
One parallel work~\cite{ren_physics-informed_2022} shares some similarities to the spatial decomposition of NeuralStagger, which leverages pixel shuffle and physics-constrained loss in the super-resolution tasks. However, we are different in target and solution. For example, we train multiple solvers to work in full parallelism and obtain the solution in multiple levels of resolution without training them again.

\section{Methodology}
\label{methodology}
\subsection{Preliminaries}
Consider a connected domain $\Omega\subseteq\mathbb{R}^n$ with boundary $\partial\Omega$, and let $(\mathcal{A},\mathcal{U},\mathcal{V})$ be separable Banach spaces. 
Then the parametric PDEs can be defined as the form
\begin{equation}\label{eq:pdes}
    \mathcal{S}(\mathbf{u},\mathbf{a})(\mathbf{x})=0, \quad \mathbf{x}\in\Omega
\end{equation}
where $\mathcal{S}:\mathcal{U}\times\mathcal{A}\rightarrow\mathcal{V}$ is a linear or nonlinear differential operator, $\mathbf{a}\in\mathcal{A}$ denotes the parameters under certain distribution $\mu$,
such as coefficient functions or boundary/initial conditions,
and $\mathbf{u}\in\mathcal{U}$ is the corresponding unknown solution function. Further, we can define the solution operator of the parametric PDE $G:\mathcal{A}\rightarrow\mathcal{U}$, which maps two infinite-dimensional function spaces. 

A main branch of works in neural PDE solvers approximate the solution operator by discretizing the functions into finite-dimensional spaces denoted by $\hat{\mathcal{A}}$ and $\hat{\mathcal{U}}$ and learning the mapping $f_\theta:\hat{\mathcal{A}}\rightarrow\hat{\mathcal{U}}$. 
Correspondingly, we have the discretized version of the PDE's operator $\mathcal{S}$ by certain finite-dimensional approximations such as the finite difference method (FDM) and finite element method (FEM), which is denoted by $\hat{\mathcal{S}}$. 
We denote the vector of the function values in a mesh with the hat symbol, e.g., $\hat{a}$ is the vector of the PDE's parameter $\mathbf{a} \sim \mu$.
Then the physics-constrained loss is defined by forcing the predicted solution $\hat{u}\in\hat{\mathcal{U}}$ to satisfy $\hat{\mathcal{S}}$ given $\hat{a}\in \hat{\mathcal{A}}$.
For example, LordNet~\citep{shi_lordnet_2022} proposed the general form with the mean squared error as follows,
\begin{equation}
    L(\theta)=\mathbb{E}_{a\sim\mu} ||\hat{\mathcal{S}}(f_{\theta}(\hat{a}), \hat{a})||^2,
\end{equation}

In this paper, we mainly focus on time-dependent problems as follows,
\begin{equation}
    \mathcal{S}(\mathbf{u},\mathbf{a})(t, \mathbf{x})=0, \quad 
    (t, \mathbf{x}) \in[0, T] \times\Omega
\end{equation}
The temporal dimension is discretized with the timestep $\Delta t$ and the neural solver solves the PDE in an auto-regressive way,
\begin{equation}
    \label{eq:auto_regressive}
    \hat{u}_{t+\Delta t} = f_{\theta}(\hat{u}_t, \hat{a})
\end{equation}
where $\hat{u}_t$ is the corresponding discretized vector of the function $\mathbf{u}$ at time $t$.
Notice that similar to traditional numerical methods, the resolution of the finite-dimensional approximation in physics-constrained loss, either in the spatial dimension or in the temporal dimension, must be sufficiently high, otherwise, the approximation error will be too large to guide the neural PDE solver. This leads to huge input and output as well as large neural networks to ensure expressiveness, whose inference would also be slow.

\subsection{NeuralStagger}
\label{sec:methodology}
We propose a general methodology called NeuralStagger to gain further accelerations by exploiting the potential parallelism in the neural PDE solver. NeuralStagger decomposes the original learning task that maps $\hat{u}_t$ to $\hat{u}_{t+\Delta t}$ into several parallelizable subtasks in both spatial and temporal dimensions. The meshes of the subtasks spread evenly in the original field and stagger with each other. Then we can handle each subtask with a computationally cheap neural network.
The decomposition strategy is introduced as follows.

\textbf{Spatial decomposition.} The upper part of Figure~\ref{fig:sp-strategy} shows the 2-dimensional example with regular mesh. 
We first split the grid into patches of the size $s_H \times s_W$ and construct a subgrid by selecting only one point in each patch, resulting in $s_H \times s_W$ subgrids evenly spread in the domain.
We denote the functions in each sub-grid as ${\hat{u}^{i,j}_t}$ and ${\hat{a}^{i,j}_t}$ where $i$ and $j$ represents the relative position of the sub-grid in horizontal and vertical directions. 
Then we use $s_H\times s_W$ neural networks to learn to predict the solution at $t+\Delta t$ as follows,
\begin{equation}
    \hat{u}^{i,j}_{t+\Delta t} = f_{\theta_{i,j}}(\hat{u}^{i,j}_t, \hat{a}^{i,j}),
\end{equation}
where $f_{\theta_{i,j}}$ is the neural network for the sub-grid at the position $(i,j)$. The outputs $\hat{u}^{i,j}_{t+\Delta t}$ compose the solution at the original grid.
Then the neural networks can be jointly trained with the physics-constrained loss defined on the original grid.
Notice that the neural networks are independent of each other and can be fully paralleled. 
As the input and output decrease by $s_H\times s_W$ times, the neural network can be much smaller and faster than the original one to be used for the neural solver.
The decomposition rules can be extended to higher-dimensional cases. 
In addition, the learning tasks at the subgrids are quite close to each other, except for the difference in the boundary of the domain, so we share the parameters of the neural networks $f_{\theta_{i,j}}$ to reduce redundancy and accelerate training. 
Meanwhile, because there are often tiny differences between the inputs of the subtasks, we encourage the neural network to distinguish them by adding positional information of each grid point as additional input channels.

\textbf{Temporal decomposition.} We can treat the temporal dimension as a 1-dimensional grid with a fixed step $\Delta t$. Thus we can also decompose the grid into $s_T$ sub-grids by selecting a point for every $s_T$ points, where instead of predicting $\hat{u}_{t+\Delta t}$, the neural network predicts $\hat{u}_{t+s_T\Delta t}$,
\begin{equation}
    \hat{u}_{t+s_T\Delta t} = f_{\theta}\left(\hat{u}_t, \hat{a}\right),
    \label{eq:temporal_decomposition}
\end{equation}
Given the solution sequence from $t$ to $t+\left(s_T-1\right)\Delta t$ denoted by $\hat{u}_{t,s_T}$ for simplicity, we can get the next sequence of the solution $\hat{u}_{t+s_T\Delta t, s_T}$. Then the physics-constrained loss is defined on the sequence with timestep $\Delta t$, as is shown in the lower part of Figure~\ref{fig:sp-strategy}. 
% \textcolor{red} {at the training stage, multi-step or single-step?}
Once the neural network is trained, we can generate the sequence $\hat{u}_{t+s_T\Delta t, s_T}$ by running the neural network inference of Formula~\ref{eq:temporal_decomposition} with $s_T$ threads in parallel with inputs $\hat{u}_{t, s_T}$. The non-auto-regressive process can generate the solution in $s_T$ time steps within one inference step, which can be much faster than the original version 
%(Figure~\ref{fig:autoreg}) 
with $s_T$ inference steps. 
Note that though we only need the initial condition for the coarsest-resolution test, we must prepare the first $s_T$ states with numerical solvers for training and the high-resolution test. However, this drawback is neglectful for long-time simulations.

The spatial and temporal decompositions are orthogonal and can be used at the same time.
We denote the joint decomposition operator as $D_{\mathbf{s}}$, the transformation operator of the neural networks as $F_{\Theta}$ and the reconstruction operator $E_{\mathbf{s}}$, where $\mathbf{s}$ represents all decomposition factors including $s_H$, $s_W$ and $s_T$, $\Theta$ represents all parameters of the neural network group. The physics-constrained loss with the spatial-temporal decomposition can be written as,
\begin{equation}
     L(\Theta)=\mathbb{E}_{\hat{u}_{t,s_T}} ||\hat{\mathcal{S}}\left(E_{\mathbf{s}} \left(	
F_{\Theta}\left(
D_{\mathbf{s}}\left(\hat{u}_{t,s_{T}}, \hat{a}\right)\right)\right), \hat{u}_{t,s_T}, \hat{a}\right)||^2.
\end{equation}

In addition, as the sub-grids spread evenly in the domain of the PDE, each of them can be seen as the down-sampled version of the original problem, where a local patch is reduced to the point at a fixed relative position in the patch. Therefore, the learned neural networks are naturally coarse-resolution solvers to the PDE. 
% With this strategy, the coarse-resolution model can be trained and the fine-resolution results can be easily and totally parallel computed. 
Suppose $(H, W, T)$ is the tuple of the original height, width, and time span that the physics-constrained loss is conducted on. Then the coarse-resolution solvers are conducted on the resolution $(\frac{H}{s_H},\frac{W}{s_W},  \frac{T}{s_T})$.
Meanwhile, we can infer multiple levels of resolutions ranging from that of coarse-resolution solvers to the original one, all of which can reach the same speed by parallelism. 

\subsection{Choice of the decomposition factors}
\label{sec:factor_choice}
Obviously, the acceleration effect by NeuralStagger grows as we use larger $s_H$, $s_W$ and $s_T$. However, these decomposition factors cannot be arbitrarily large. We conclude two potential constraints, i.e., the increased complexity of the learning task and the information loss in the input. We would like to leverage the following 2-dimensional diffusion equation with the periodic boundary condition as an example to explain the two constraints,
\begin{align}
\label{eq:diffusion}
    \frac{\partial u(x, y, t)}{\partial t} &=\Delta u(x, y, t), &x, y, t \in [0, 1], \\
% & \partial_t u\left(x, y, t\right)=\Delta u\left(x, y, t\right), x\in \Omega, \\
    u(x, y, 0) &= f(x, y), & x, y\in [0, 1],
\end{align}
where $u$ is the density function of diffusing material, $\Delta$ is the Laplacian operator and $f$ is the function of the initial condition. 
We use the regular mesh with $d$ points in total and leverage the central difference scheme with the spatial step $\Delta x$ and temporal step $\Delta t$. Then the PDE is transformed into a matrix equation on the discretized solution at a certain time $t$, denoted by $\hat{u}_{t} \in \mathbb{R}^d$. 

\begin{figure}
\centering
\includegraphics[width=0.4\textwidth]{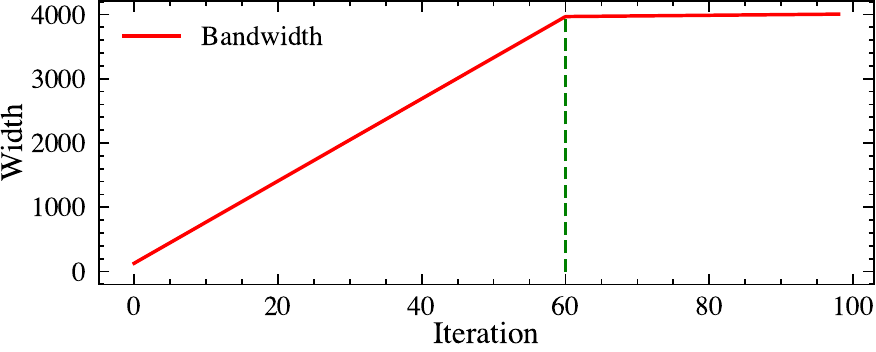}
    \caption{The bandwidth curve.}
    \label{fig:temporal-diffu}
\end{figure}

\textbf{Increased complexity of learning task.} 
\label{sec:tsf}
For the temporal dimension, we find that the larger decomposition factor might make the mapping from the input to the prediction more complex. 
%Here we take the 2-D diffusion equation with a constant diffusion coefficient D as an example:
%Taking the diffusion equation as an example, we approximate the derivatives by FDM, so the PDE turns to a matrix equation on the discretized solution function $\hat{u}_t$. 
For the linear diffusion equation, we can explicitly calculate the transfer matrix from $\hat{u}_i$ to $\hat{u}_{i+\Delta t}$ based on the matrix equation.
Suppose the transfer matrix is $T_i \in \mathbb{R}^{d\times d}$.
By iterative applying the transfer matrix, we can get the transformation from the initial condition $\hat{u}_0$ to the solution at any time step $k$ as follows,
\begin{equation}
    \label{eq:diffusion_derivatives}
    \hat{u}_{k\Delta t} =  \hat{u}_0 \prod_{0}^{k-1}{T_i}.
\end{equation}
For notational simplicity, we denote the resulting transfer matrix 
from $\hat{u}_0$ to $\hat{u}_{k\Delta t}$ as $\mathcal{T}_{k}$.
By certain arrangements, $\mathcal{T}_{k}$ is a band matrix where the non-zero values are centralized around the diagonal. The bandwidth indicates the sparsity of the matrix as well as how local the points in the mesh entangle with each other.
We observe that the bandwidth grows linearly with regard to $k$. 
For example, Figure~\ref{fig:temporal-diffu} shows the case of $d=64^2$.  
When the $k \geq 60$, the matrix is dense and every element in $\hat{u}_{k\Delta t}$ is a weighted summation of almost all the elements in $\hat{u}_t$. 
This indicates that increasing $k$ may make the entanglements between the grid points more complex, leading to a harder learning task for the neural network.

\textbf{Information loss.}
By spatial decomposition, each subgrid only reserves a small part of the original grid.
Obviously, it may introduce the problem of information loss if the dropped points are important for the prediction in the subtasks. 
Here we theoretically characterize the information loss caused by spatial decomposition under the linear model setting, i.e., $f(\hat{u}_t) = \hat{u}_t W^*$. 
Consider the diffusion equation and the corresponding matrix equation. With some abuse of notation, the superscript $i$ denotes the index of training samples, such as $\hat{u}_t^i$ and the bold symbol without the superscript $i$ denotes the matrix composed of all the samples, such as $\hat{\boldsymbol{u}}_t$. With $N$ training samples, the physics-constrained loss aims to learn the parameters $W^*$ of the linear model that satisfies:
\begin{align}\label{eq_min}
    W^*=\argmin_{W} \frac{1}{N}\sum_{i=1}^N\|\hat{u}^i_{t}W- y^i\|^2,
\end{align}
where $y^i$ denotes the rest parts of the matrix equation. 
By applying spatial decomposition, the input and output are equally partitioned into $K=s_Hs_W$ subgrids $\{\hat{u}_t^{1},\cdots,\hat{u}_t^{K}\}$ and $\{\hat{u}_{t+1}^{1},\cdots,\hat{u}_{t+1}^{K}\}$. Then according to the physics-constrained loss, the optimization goal becomes:
\begin{align}\label{eq_decompmin}
    W_1^*,\cdots,W_K^*=\argmin_{W_1,\cdots,W_K}\frac{1}{N}\sum_{i=1}^N\sum_{k=1}^K\|(\hat{u}_t^{i,k}W_k-y^{i,k})\|^2,
\end{align}where $W_k\in\mathbb{R}^{m\times m}, m=d/K$ for $k=1,\cdots,K$. The next proposition shows a sufficient condition for equal prediction for Eq.(\ref{eq_min}) and Eq.(\ref{eq_decompmin}).
\begin{proposition}
\label{prop}
If  
$rank(\hat{\boldsymbol{u}}_t)=rank(\hat{\boldsymbol{u}}^k_t)$, the model $\hat{\boldsymbol{u}}_tW^*$ and $\hat{\boldsymbol{u}}_t^{k}W^*_k$ will make the same prediction on $\boldsymbol{y}^k$.
\end{proposition}
We put the proof in Appendix~\ref{sec:inf_loss_sd}. 
In practice, the proposition is held approximation in many physical scenarios. This is because local patches of size $s_Hs_W$ do not distribute arbitrarily in the ambient space $\mathbb{R}^{s_Hs_W}$, but rather live in some low-dimensional manifold. Hence, there is much information redundancy in $\hat{\boldsymbol{u}}_t$ and with careful settings of $s_H$ and $s_W$, the rank after the decomposition does not change much.

In addition, the information loss can be made up by adding information that describes the local patches to the input. Such supplementary information can be extracted by either neural network layers or feature engineering, which can be designed for specific problems. 
In Section~\ref{sec:feature-info-loss}, we test several choices of design for fluid dynamics systems.

\section{Experiments}
\label{experiments}
To evaluate the acceleration effect and accuracy of the proposed method, we test three cases of fluid dynamics simulation governed by the Navier-Stokes equation. 
We first target two benchmark settings, i.e., the periodic boundary condition and the lid-driven cavity boundary condition~\cite{Zienki2006}. 
In both settings, the initial condition changes, and the neural PDE solver learns to generalize to various initial conditions. 
Next, we test the more challenging case called flow around obstacles in both the 2-dimensional (30 thousand grid points) and 3-dimensional ($\sim$0.5 million grid points) cases. 
The neural PDE solver is trained to generalize to different obstacles as well as initial conditions.
Thirdly, we evaluate the capability of the learned solvers to handle the inverse problem.
At last, we also demonstrate that adding supplementary information to the input helps alleviate the problem of information loss. 
\begin{figure*}[t]
     \centering
     \includegraphics[width=0.95\textwidth]{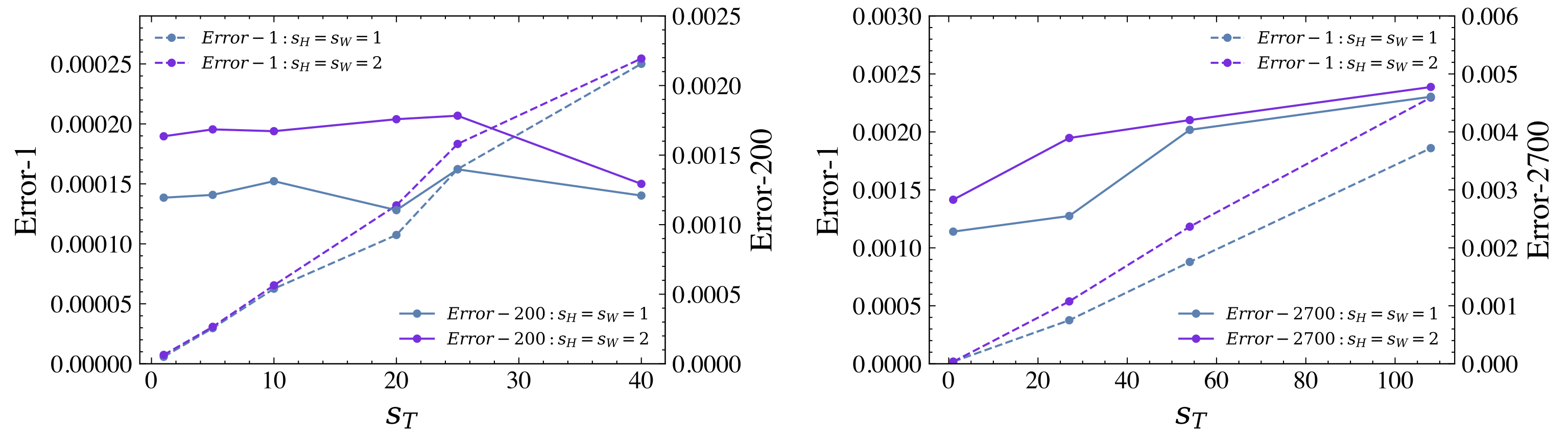}
     %\vspace{-12pt}
    \caption{Tests on Navier-Stokes equation with (left) periodic boundary condition and (right)  Lid-driven cavity boundary condition.}
    \label{fig:exp}
\end{figure*}

In general, we consider the incompressible Navier-Stokes equation as follows:
\begin{align}
\label{eq:nse-ori}
\rho\left(\frac{\partial \vec{v}}{\partial t}+(\vec{v} \cdot \nabla) \vec{v}\right) &=-\nabla p+\mu \Delta \vec{v}+\vec{f}\\
\nabla \cdot \vec{v} &=0 
\end{align}
where $\vec{v}$ is the fluid velocity field, $p$ is the pressure field, $\mu$ is the viscosity, and $\vec{f}$ is the external force. 
In all experiments, we trained neural networks with Adam optimizer and decayed learning rates.
The speed-up effect is mainly evaluated with the computational load called GMACs (Giga multiply-accumulate operations) per GPU card rather than the inference time of the neural network because it depends largely on the computational hardware.
See Appendix~\ref{sec:exp_detail} for more details on the implementation.

\subsection{Periodic and Lid-driven cavity boundary condition} 
\label{sec:nse_lid}
\begin{figure*} 
 \centering
 \includegraphics[width=0.78\textwidth]{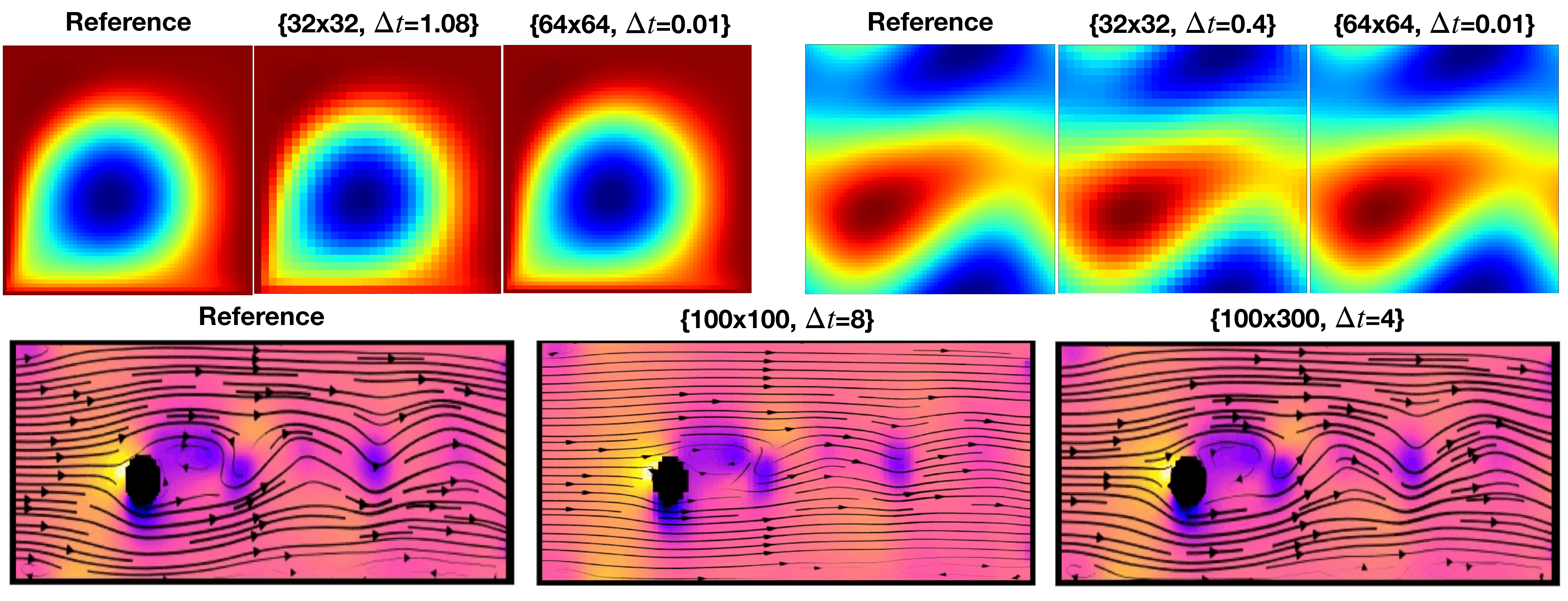}
  \vspace{-11pt}
  \caption{The predictions in two resolutions. \textbf{Top}: lid-driven cavity boundary condition (left) and periodic boundary condition (right) and \textbf{Bottom}: flow around obstacles. }
  \label{fig:vis_all}
\end{figure*}

We first test the Navier-Stokes equation with the periodic boundary condition and the lid-driven cavity boundary condition.
In both cases, the physics-constrained loss is obtained by discretizing the vorticity-stream equation with the central-difference scheme and the Crank-Nicolson method in the $64\times 64$ regular mesh. The time step $\Delta t$ is $1\mathrm{e}-2$ and the viscosity $\nu$ is $1\mathrm{e}-3$. 
We use the popular FNO~\citep{li_fourier_2020} to test the accuracy and speed in different settings of decomposition factors.
The ground truth is obtained by FDM. 
We evaluate the accuracy by auto-regressively running the inference of the neural solver across the target length along time $L_T$ and compare the terminal state with that from the ground truth. 
Note that we compare all the results on the original mesh and thus the spatially decomposed results reconstruct to the $64\times 64$ resolution for evaluation.
We measure with the relative error which is calculated by dividing the L2 norm of the error by the L2 norm of the ground truth. 
The measurement is denoted by Error-$k$ where $k$ is the number of time steps. 
Following the notations in Section~\ref{sec:methodology}, the decomposition factors along $x$ dimension, $z$ dimension, and the temporal dimension are denoted by $s_W$, $s_H$, and $s_T$. 
In general, NeuralStagger achieves acceleration in both cases without losing much accuracy. As you can see in Figure~\ref{fig:vis_all}, the coarse-resolution solver is also accurate when applied alone without reconstruction.

In the case of the periodic boundary condition, the target length along time $L_T$ equals 2, which is 200 time steps.
The flow is driven by the external force $\vec{f}$, which is introduced in Appendix~\ref{sec:exp_detail}.
As you can see in Figure~\ref{fig:exp} (left), the relative errors of the learned neural solvers are lower than 0.2\% in all settings of spatial and temporal decomposition factors. 
In terms of speed, with the most aggressive setting $s_T=40, s_H=s_W=2$, and full parallelism, 
the GMACs of the 200-time-steps inference decrease from $31.92$ to $0.24$, which is $133$ fold reduction, corresponding to $47\times$ speed-up in time if the inference is conducted on A100 cards. 
We can also observe some trends in accuracy with regard to the choice of spatial and temporal factors.
Error-1 grows like a linear function with the temporal factor $s_T$ in both spatial factor settings. The reason is that the learning task becomes more complex as we discuss in Section~\ref{sec:factor_choice}, and with the neural network unchanged, the accuracy drops.
Meanwhile, the accumulated errors, i.e., Error-200, almost keep at the same level. This is because the steps in the auto-regressive procedure reduce as $s_T$ grows, e.g., when $s_T=40$, the neural networks for subtasks only predict $200/40=5$ steps ahead. The benefit perfectly neutralizes the detriment of the increased task complexity. 

In the case of the lid-driven cavity boundary condition, the fluid acts in a cavity consisting of three rigid walls with no-slip conditions and a lid moving with a steady tangential velocity 1.
We set the length of time $L_T=27$, much larger than that with the periodic boundary, to see if the simulation converges to the right steady state. 
With larger $L_T$, we try larger temporal skip factors such as $s_T=108$.
As is shown in Figure~\ref{fig:exp} (right), the relative errors are all controlled below 0.5\% even after 2700 time steps. 
Again, with the most aggressive setting $s_T=108, s_H=s_W=2$ and full parallelism, 
the GMACs per card of 2700-steps inference decreases from $404.92$ to $1.12$, which is $362$ fold reduction, corresponding to $119\times$ speed-up in time with A100 cards.
Different from the periodic boundary condition, the accuracy drops when we increase $s_T$, because in this case, the increase of $s_T$ brings more detriments of task complexity than the benefits from the shorter auto-regressive sequence. 

\subsection{Flow around obstacles}
\label{sec:flow-around}
In this section, we evaluate NeuralStagger in a larger and more complex setting called flow around obstacles. 
The setting is the same as that used in~\cite{Wandel2020}, which is also our baseline. 
The fluid runs through a pipe, where we put different shapes of obstacles to affect the flow, including rotating cylinders and walls constructing a folded pipe. The external forces in Eq.~\ref{eq:nse-ori} are neglected and set to 0. 
The neural solver is trained to generalize to different settings of the obstacles, including the shape and the velocity on the surface as well as the inflow/outflow velocities. Then we evaluate the neural solver in 5 randomly sampled configurations in both the cylinder case and the folded pipe case. You may refer to Appendix~\ref{sec:exp_detail} for more details.
We leverage the same configurations as that in~\citet{Wandel2020} including the discretization method, the physics-constrained loss, training strategies, the input features, the predicted variables as well as the evaluation metric.
Specifically, the rectangular domain is discretized into a $100 \times 300$ regular mesh and $\Delta t=4$.
The physics-constrained loss is used as the evaluation metric, measuring to what extent the prediction at the next time step satisfies the PDE given the current fluid state and the boundary conditions. 
As the fields of the fluid change much over time, we maintain a training pool initialized with a set of initial conditions and incrementally enrich it as the training goes on. This is achieved because the predictions from the neural network can be seen as new data if the neural network has been well-fitted in the current pool. One can refer to \citet{Wandel2020} for more details.

\citet{Wandel2020} leverages U-net as the neural solver, but to demonstrate the full potential of NeuralStagger,
we also try the other two neural network architectures, i.e., FNO and LordNet ~\citep{shi_lordnet_2022} which also leverages the physics-constrained loss to train the neural PDE solver.
The experiments in Table~\ref{tab:fap_exp} show that LordNet outperforms the other two neural networks in the baseline setting without NeuralStagger. Therefore, we use LordNet for further experiments on the choice of spatial and temporal factors (see the performance on other networks in Appendix~\ref{app:fulllist}).
We find that in this case, the information from the $100 \times 100$ grid ($s_H=1, s_W=3$) is sufficient to achieve comparable results to the U-net baseline, while larger spatial steps will introduce too much information loss.
In addition, we observe that increasing the temporal factors hurts the accuracy more obviously than those in the periodic boundary condition and the lid-driven boundary condition, though the accuracy is still comparable to U-net even with $s_T=16$.
We believe this is because the dataset is incrementally explored by maintaining a training pool and enriching it with the neural network's predictions during training while those predictions may not be accurate. 
As the physics constrained loss is defined on $\hat{u}_{t+(s_T-1)\Delta t}$ and $\hat{u}_{t+s_T\Delta t}$, inaccurate $\hat{u}_{t+(s_T-1)\Delta t}$ may mislead the neural network to the wrong direction.
When we increase $s_T$, more errors will be accumulated along the sequence from $\hat{u}_{t}$ the $\hat{u}_{t+(s_T-1)\Delta t}$ and the training will be harder. Designing training algorithms to support NeuralStagger better remains unexplored and we leave it for future work.

In terms of speed, the choices of spatial and temporal factors lead to different levels of acceleration, as is shown in Table~\ref{tab:fap_exp}, where GMACs per card is the average computational load of simulation for 16 timesteps.
As you can see, for LordNet, there is an approximately linear relationship between GMACs per card and the inverse of each decomposition factor.
The largest factor configuration to keep the accuracy comparable to the baseline is $s_T=16, s_H=1, s_W=3$, leading to the largest decrease in GMACs per card, i.e., $1/58$ of LordNet without NeuralStagger, which can be attributed to $1/16$ from the decrease of temporal steps and approximately $1/4$ from the decrease of the neural network size and input size. 
The actual speed-up effect depends on the hardware devices we use. For example, when tested with NVIDIA RTX 2080ti, it leads to $23\times$ speed-up and when tested with A100, it leads to $17\times$ speed-up as the baseline of LordNet is already very fast.
\begin{table}[!htb]
    \centering
    \vspace{-0.2cm}
    \caption{
The Performance of NeuralStagger with different decomposition factors and neural networks in the flow-around-obstacles settings. 
}
    \setlength{\tabcolsep}{0.011\columnwidth}
    \begin{tabular}{c|c|c|c|c|c}
         \toprule
        \multirow{2}{*}{Config} &
         Temporal & 
         Spatial &
         Folded & \multirow{2}{*}{Cylinder} & GMACs\\
         &
         factor & factors & pipe &  & per card\\
         \midrule
         U-net & - & - & 6.32 e-5 & 1.24 e-4 & 29.60\\
         \hline
         FNO & - & - &4.01 e-4  & 4.54 e-4 & 18.51 \\
        \hline
         \multirow{7}{*}{LordNet}&
         - & - & 1.05 e-5 & 4.11 e-5  & 71.04\\
         &
         1 & (1, 3) & 2.21 e-5 & 8.97 e-5 & 19.84\\
         &
         1 & (2, 6) & 5.00 e-4 & 2.98 e-3 & 4.46 \\
         &
         2 & (1, 1) & 3.59 e-5 & 7.68 e-5 & 35.52 \\
         &
         2 & (1, 3) & 5.51 e-5 & 1.19 e-4 & 9.92\\
         &
         8 & (1, 3)  & 3.93 e-4   & 6.60 e-4 & 2.48\\
         &
         16 & (1, 3)& 3.47 e-4  & 8.55 e-4 & 1.24 \\
        \bottomrule
    \end{tabular}
    \label{tab:fap_exp}
\end{table}

\textbf{Accelerations in the 3D case}.
We would like to further stress that the acceleration effect by the spatial decomposition is even larger in 3-dimensional scenarios. 
Following the work~\cite{wandel_teaching_2021}, we test the 3-dimensional flow around obstacles case with U-net3d, FNO3d, and LordNet3d.
In this case, the rectangular domain is discretized into $128\times64\times64$ regular mesh and $\Delta t$=4. We still train the neural network for the fluid dynamics from scratch like what we have done in 2-dimensional cases. More details and results can be found in Appendix~\ref{sec:exp_detail}.  
We evaluate the choice of decomposition factors per dimension. As you can see in Table~\ref{tab:3d-fao}, without losing much in accuracy, i.e., keeping the PDE residuals in the same magnitude, the spatial decomposition alone can introduce up to about $19$ fold decrease on GMACs. What's more, the experiments demonstrate that for more realistic and challenging tasks, e.g. learning 3D fluid dynamics, the NeuralStagger can still work well.
\begin{table}[!htb]
\label{tab:3d-fao}
    \centering
    \vspace{-0.1cm}
    \caption{The performance of NeuralStagger with different neural networks in the 3-dimensional flow-around-obstacles setting.}
    \setlength{\tabcolsep}{0.042\columnwidth}
    \begin{tabular}{c|c|c|c}
         \toprule
         \multirow{2}{*}{Network} & Spatial & PDE & GMACs  
         \\
         & factors & residual & per card \\
        \midrule
         \multirow{1}{*}{U-net3d} & - & 1.05 e-4 & 62.73\\
         \hline
         \multirow{1}{*}{FNO3d} & - & 1.15 e-4 & 21.89\\
         \hline
         \multirow{3}{*}{LordNet3d} & - & 1.01 e-4 & 73.86 \\
         & (2,2,2) & 4.54 e-4 & 8.23\\
         & (4,2,2) &4.67 e-4 & 3.99\\
        \bottomrule
    \end{tabular}
    \label{tab:flow_ap_3d}
\end{table}

\subsection{Application in optimal control}
To further showcase the capability of the neural solver with NeuralStagger on the inverse problem, we
conduct the optimal control experiment introduced in \citet{Wandel2020}. 
The task is to change the flow speed to control the shedding frequency of a Kármán vortex street behind an obstacle. 
Here, we take an example of LordNet using NeuralStagger with setting $s_H=1, s_W=3, s_T=2$, which outperforms the baseline U-net in Table~\ref{tab:fap_exp}. 
We observe that it tackles this inverse problem much faster and also stabler than the baseline does.
One may refer to Appendix~\ref{app:freq_control} and Figure~\ref{fig:frequency_control} for more details about the settings and results.

\subsection{Supplementary to the information loss}
\label{sec:feature-info-loss}
As is discussed in Section~\ref{sec:factor_choice}, introducing supplement information can alleviate the bad influence of spatial decomposition on accuracy. We design and try two methods for the fluid cases. The first is the vorticity field that describes the local spinning motion of the fluid. 
While it introduces an additional input channel, the computational overhead does not increase much as we only change the first layer to fit the input size.
The second is to add positional encoding (PE) that embeds the coordinates in the original grid to each of the input channels so as to help distinguish different sub-tasks.
One may refer to Appendix~\ref{app:feature_loss_imp} for more details.
As is shown in Table~\ref{tab:flow_ap_2d_feature}, with either choice, we observe obvious performance gains in the flow around obstacles case. 
\begin{table}[!htb]
    \centering
    \vspace{-0.1cm}
    \caption{The performance of U-net w/wo the supplement information in flow-around-obstacles setting.}
    \setlength{\tabcolsep}{0.006\columnwidth}
    \begin{tabular}{c|c|c|c|c|c|c}
         \toprule
         Spatial  & \multicolumn{3}{c}{Folded pipe} \vline & \multicolumn{3}{c}{Cylinder}  
         \\
          factors& None & Vorticity & PE & None & Vorticity & PE \\
        \midrule
         (1,3) & 1.94e-4 & 1.79e-4 & 8.06e-5 &2.84e-4 & 2.56e-4 & 1.46e-4\\
         (2,6) & 4.35e-4 & 2.28e-4  & 2.93e-4 &1.09e-3&6.21e-4& 7.46e-4\\
        \bottomrule
    \end{tabular}
    \label{tab:flow_ap_2d_feature}
\end{table}

\section{Conclusion and Limitation}
\label{conclusion}
We present NeuralStagger, a general framework for accelerating the neural PDE solver trained by physics-constrained loss.
By spatially and temporally decomposing the learning task and training multiple lightweight neural networks, the neural solver is better paralleled and much faster with sufficient computational resources.
In addition, each lightweight neural network is naturally a coarse-resolution solver and they bring the flexibility of producing the solutions on multiple levels of resolution, which is important for balancing the resolution and computational resources.
We discuss the choice of decomposition factors and empirically test their influence on accuracy and speed. The experiments in fluid dynamics simulation show that NeuralStagger brings an additional 10 to 100$\times$ speed-up over SOTA neural PDE solvers with mild sacrifice on accuracy.

One limitation of our work is that we only define the spatial decomposition over regular meshes, while it turns to the non-trivial vertex coloring problem for irregular meshes. Heuristic coloring algorithms would be useful for this problem, and we would like to explore it in future works.

\section*{Acknowledgments}
\label{ack}
The authors thank the
reviewers and area chairs for their helpful suggestions.
Xinquan Huang acknowledges financial support from King Abdullah University of Science and Technology (KAUST).

\bibliography{references}
\bibliographystyle{icml2023}
\newpage
\appendix
\section{Appendix}
\subsection{Information loss caused by spatial decomposition}
\label{sec:inf_loss_sd}
In this section, we provide the proof to proposition~\ref{prop} in the linear model setting.
In this section, we will theoretically characterize the information loss caused by spatial decomposition under the linear model setting. Note that the proof is done on the 1-dimensional diffusion equation with the explicit method for ease of understanding, but as we will see, the conclusion is the same in the case with 2 dimensions or the implicit method. 

We consider a simple 1d partial differential equation with Dirichlet boundary condition:
\begin{align}
    &\partial_t u=\Delta u, x\in \Omega\\
    &u_t(x)=f_t(x), x\in\partial \Omega
  %  &\partial_x u(x)=g(x), x\in\partial\Omega
\end{align}
Discretizing the function $u$ on grid $(x_1,\cdots,x_d)$, we denote $\hat{u}^j=u(x_j)$. We consider the finite difference discretization:
\begin{align}
    &\frac{\hat{u}_{t+1}^j-\hat{u}_t^j}{\delta t}= \frac{(\hat{u}_{t}^{j+1}-\hat{u}_{t}^{j})-(\hat{u}_{t}^{j}-\hat{u}_{t}^{j-1})}{\delta x^2}, x_j\neq \{x_1, x_d\}\\
    &\hat{u}_{t+1}^{j}=f_{t+1}(x_j), x_j=\{x_1,x_d\}
\end{align}

Given the input $\hat{u}_t\in\mathbb{R}^d$ and output $\hat{u}_{t+\Delta t}\in\mathbb{R}^d$, the output $\hat{u}_{t+\Delta t}$ is parameterized by linear model as $\hat{u}_{t+\Delta t}=\hat{u}_{t}W$ where $W\in\mathbb{R}^{d\times d}$ denotes the learned parameters. The physics constrained loss aims to learn the parameters $W^*$ of the linear model that satisfies:
\begin{align}\label{app:eq_min}
    W^*=\argmin_{W} \frac{1}{N}\sum_{i=1}^N\|\hat{u}^i_{t}W-y^i\|^2,
\end{align}where $i$ denotes the index of training samples and $y^j=f_{t+1}(x_j), x_j=\{x_1,x_d\}$; $y^j=\hat{u}_{t}^{j}-\frac{\delta t}{\delta x^2}\left((\hat{u}_{t}^{j+1}-\hat{u}_{t}^{j})-(\hat{u}_{t}^{j}-\hat{u}_{t}^{j-1})\right), x_j\neq \{x_1, x_d\}$. 

%After calculation, the minimizer for Eqn.(\ref{eq_min}) is 
%\begin{align}
%    W^*=\frac{\sum_{i=1}^N\bf({u}_i^t)^{\tau}\bf{y}_i}{\sum_{i=1}^N (\bf{u}_i^t)^{\tau}\bf{u}_i^{t}}
%\end{align}

By applying spatial decomposition, the input and output are equally partitioned into $K$ blocks $\{\hat{u}_{t}^{1},\cdots,\hat{u}_{t}^{K}\}$ and $\{\hat{u}_{t+\Delta t}^{1},\cdots,\hat{u}_{t+\Delta t}^{K}\}$. Each block contains $d/K$ coordinates
% as $\hat{u}_{t}^{k}=[\hat{u}_{t}^{k},\hat{u}_{t}^{k+d/K},\cdots, \hat{u}_{t}^{k+d/K}]$. 
Then according to the MSR loss, the optimization goal becomes:
\begin{align}\label{app:eq_decompmin}
    W_1^*,\cdots,W_K^*=\argmin_{W_1,\cdots,W_K}\frac{1}{N}\sum_{i=1}^N\sum_{k=1}^K\|(\hat{u}_t^{i,k}W_k-y^{i,k})\|^2,
\end{align}where $W_k\in\mathbb{R}^{m\times m}, m=d/K$ for $k=1,\cdots,K$.
%The minimizer for Eqn.(\ref{eq_decompmin}) is
%\begin{align}
%    \hat{W}_k^*=\frac{\sum_{i=1}^N\bf({u}_i^{t,k})^{\tau}\bf{y}_{i}^k}{\sum_{i=1}^N (\bf{u}_i^{t,k})^{\tau}\bf{u}_i^{t,k}}.
%\end{align}

%We concat $W_1^*,\cdots, W_K^*$ to be the diagonal block of a $d\times d$ matrix $\hat{W}$, i.e., $\hat{W}=diag\{W_1^*,\cdots, W_K^*\}$, then we have the following proposition.

%\begin{proposition}
%If $rank(\sum_{i=1}^N(\hat{u}^i_t)^{\tau}\hat{u}^i_t)=rank(\sum_{i=1}^N(\hat{u}_t^{i,k})^{\tau}\hat{u}_t^{i,k})$, the model $\hat{u}_tW^*$ and $\hat{u}_{t}^{k}W^*_k$ will make the same prediction on $y^k$.  
%\end{proposition}

\textbf{Proof:} We first consider the case that %$rank(\sum_{i=1}^N(\textbf{u}_i^t)^{\tau}\textbf{u}_i^t)=rank(\sum_{i=1}^N(\textbf{u}_i^{t,k})^{\tau}\textbf{u}_i^{t,k})=d/[K]$, which means 
$\sum_{i=1}^N(\hat{u}_t^{i,k})^{\tau}\hat{u}_t^{i,k}$ is full rank. The minimizer of Eq.(\ref{app:eq_decompmin}) is ${W}_k^*=(\sum_{i=1}^N (\hat{u}_t^{i,k})^{\tau}\hat{u}_t^{i,k})^{-1}({\sum_{i=1}^N(\hat{u}_t^{i,k})^{\tau}{y}_{i,k}})$. We denote the matrix $A=(\sum_{i=1}^N (\hat{u}_t^{i,k})^{\tau}\hat{u}_t^{i,k})^{-1}$, We construct a $d\times d$ matrix $B$ by letting $B(k+id/K,k+jd/K)=A(i,j),$ for $ i=0,\cdots,d/K; j=0,\cdots, d/K$; otherwise, $B(i,j)=0$. 
Then it is easy to check that the matrix $B$ is the pseudo-inverse of $\sum_{i=1}^N(\hat{u}^i_t)^{\tau}\hat{u}^i_t$. The minimizer of Eq.(\ref{app:eq_min}) is \citep{bartlett2020benign} $B({\sum_{i=1}^N(\hat{u}_i^{t})^{\tau}{y}_{i}})$. As the matrix $B$ only has non-zero values on the coordinates that correspond to the $k$-th block, we have the $k$-the block of $W^*$ equals $W_k^*$ and other blocks equal zero matrices. 
Denoting the matrix composed of all the samples with the bold symbol without the superscript $i$ such as $\hat{\boldsymbol{u}}_t$ for $\left\{ \hat{u}_t^{i} \right\}$ and $\hat{\boldsymbol{u}}^k_t$ for $\left\{ \hat{u}_t^{i,k} \right\}$, we have $\sum_{i=1}^N(\hat{u}_t^{i,k})^{\tau}\hat{u}_t^{i,k}=(\hat{\boldsymbol{u}}^k_t)^{\tau}\hat{\boldsymbol{u}}^k_t$ and $\sum_{i=1}^N(\hat{u}_t^{i})^{\tau}\hat{u}_t^{i}=(\hat{\boldsymbol{u}}_t)^{\tau}\hat{\boldsymbol{u}}_t$. By Rank–nullity theorem, it is easy to see that  $rank((\hat{\boldsymbol{u}}_t)^{\tau}\hat{\boldsymbol{u}}_t)=rank(\hat{\boldsymbol{u}}_t)$ and $rank((\hat{\boldsymbol{u}}^k_t)^{\tau}\hat{\boldsymbol{u}}^k_t)=rank(\hat{\boldsymbol{u}}^k_t)$.
Then we get the results in the proposition.

For the case that $\sum_{i=1}^N(\hat{u}_i^{t,k})^{\tau}\hat{u}_i^{t,k}\leq d/K$, we can select its maximal linearly independent group to obtain its pseudo-inverse and apply similar analyses to get the results. In the case of the implicit method, the term $\hat{u}^i_{t}W$ in the physics constrained loss becomes $\hat{u}^i_{t}WV$ where $V$ is an invertible matrix. This also does not change the conclusion.

\subsection{Implementation details}
\label{sec:exp_detail}
We implemented FNO with the original 2-dimensional version in the official repository, where we set the truncation mode to 12 and the width to 64.
For the LordNet, we only stack 2 Lord modules and fix the channel count to 64 in all layers. 
In the position-wise embedding of the 2 Lord modules, we stack two 1$\times$1 Convolutional layers, where the hidden embedding contains 256 and 128 channels separately, and GELU activation is used between the Convolutional layers. 
The implementation of Unet is based on the U-Net architecture \citep{ronneberger2015u} with 20 hidden channels, which is consistent with that in \citep{Wandel2020}
The learning rates and training samples are described as follows. To keep out the potential influence of computational resources like cores and memory, we test the speed of NeuralStagger under the setting that each coarse-resolution solvers have sufficient resources to use. 
Therefore, we run each solver on Nvidia A100 GPUs with the batch size equals to 1. The time per step shown in Table~\ref{tab:fap_exp} is calculated by dividing the inference time of the coarse-resolution solver by the temporal factor $s_T$. The time of decomposition and reconstruction is ignored because the operation supported by `pixel shuffle' is super efficient. 
We also calculated  GMACs (Giga multiply-accumulate Operations) per card, which is the average computational load of simulation for 16 timesteps. Note that for the GMACs of FFT operation, we calculate it by 2$Nlog_2N$, where N is the number of spatial grids.

\textbf{Periodic Boundary Condition} 
We generate the data with random fields to generate a periodic function on a 64$\times$64 grid with a time-step of 1e-2 where we record the solution every time step, where the external force is fixed $f(x)=0.1sin(2\pi(x+y)) + cos(2\pi(x+y))$. 
For the perioidc boundary and lid-driven boundary conditions, we use the vorticity-stream function form of  Eq.~\ref{eq:nse-ori} as the physics-constrained loss. With the Helmholtz decomposition to Eq.~\ref{eq:nse-ori}, we rewrite the Navier-Stokes equation:
\begin{align}\label{eq:nse}
    \frac{\partial\omega}{\partial t} = -\frac{\partial \psi}{\partial y}\frac{\partial \omega}{\partial x} + \frac{\partial \psi}{\partial x}\frac{\partial \omega}{\partial y} &+ \frac{1}{\text{Re}}\left(\frac{\partial^2 \omega}{\partial x^2} + \frac{\partial^2 \omega}{\partial y^2}\right) \\ \label{eq:nse1}% , t\in (0, T] \\
        \frac{\partial^2 \psi}{\partial x^2} + \frac{\partial^2 \psi}{\partial y^2} &= - \omega  ,
\end{align}
\label{eq:nse2}
where $\omega$ is the vorticity function, $\psi$ is the stream function, and $\text{Re}$ is the Reynolds number. 
The initial condition $\omega_0$ is generated by random field satisfying the distribution $\mathcal{N}\left(0,8^{3}(-\Delta+64 I)^{-4.0}\right)$.
We use 6000 states for training. 
In this case, we use FNO to test NeuralStagger and decay the initial learning rate 3e-3 with a factor of 0.9 every 5000 iterations.

\textbf{Lid-driven Cavity boundary condition}
We generate data on a 64$\times$64 domain but we train the neural network to predict the values of $\psi$ inside the boundary, which is a 2-dimensional matrix of the shape $(H-2)\times(W-2)$. 
The random initial conditions are generated in the same way as the periodic boundary conditions. To make the initial state consistent with the boundary condition, we solve with the numerical solver for the first $T_0=1.98$ and use $\omega_{T_0}$ as the initial state.
We use 8200 states for training with FNO, and decay the initial learning rate 3e-3 with a factor of 0.9 every 10000 iterations. 

\textbf{Flow around Obstacles} 
The data generation is the same as the setting used in \citep{Wandel2020}, where the resolution of the domain is 100$
\times$300, $\Delta t=4, \rho=4, \mu=0.1$. In training, different types of environments are used including magnus, box, and pipe. The locations and the velocity are variable during the training, e.g., the velocity is ranged from 0.0 to 3 m/s, the diameter of the obstacle is ranged from 10 to 40, and the coordinate x of the location is randomly from 65 to 75 and the coordinate y of that is from 40 to 60. 
And then for the test, we randomly select the location and flow velocity to test and in our experiment, the Reynolds number of tests is 517. 
In this case, we train the model from scratch without any data for $s_T=1$. 
For $s_T>1$, we use the benchmark to pre-generate the initial sequence $\hat{u}_{0,s_T}$ for training. 
During the training, the predicted samples $\hat{u}_{s_T,s_T}$ are not accurate, and feeding them into the training pool might collapse the training.
Thus, we use the benchmark to correct the parts of predicted samples $\hat{u}_{s_T+1,s_T-1}$. 
The learning rate is 1e-3 for Lordnet and 3e-3 for FNO, both with a factor of 0.9 every 5000 iterations. The quantitative comparison in this paper is conducted on a 100$\times$300 grid. For the optimal control of the vortex shedding experiment, the domain size is 100$\times$300, and used the trained neural PDE solver based on the above training settings. The Reynolds number here is 880. The optimizer for both Unet and LordNet is Adam optimizer with a learning rate of 1e-3.

\textbf{3-dimensional Flow around obstacles} Similar to the 2-dimensional case, the training and testing were conducted in a 128$\times$64$\times$64 domain with $\Delta t$=4 and the neural networks are trained from scratch. In training, various types of environments are used in which including obstacles such as boxes, spinning balls, or cylinders. The locations of the obstacles and inflow/outflow velocities are variable, e.g., the velocity is ranged from 0.0 to 3.0 m/s, the diameter of the obstacle is ranged from 10 to 45, and the coordinates, y and z, of the location is randomly from 22 to 42 and the coordinate x of that is from 22 to 42 and 86 to 106. And then for the test, we choose the same benchmark test in \cite{wandel_teaching_2021} to compare the performance of different neural networks and different spatial-temporal factors. We use a learning rate of 1e-3 and Adam optimizer to train the model. The networks used are extensions of the original networks to the 3D version. The U-net3d (baseline) is the version from the repository of \cite{wandel_teaching_2021}. For the FNO3d, we set the truncation mode to 12 and the width to 64. For the LordNet3d, we only stack 2 Lord modules and fix the channel count to 64 in all layers. 
The quantitative comparison is based on the PDE residuals on the 128$\times$64$\times$64 domain. Here, the GMACs are the cost for one timestep of simulation.

\textbf{Difference scheme} We use the 2nd-order central finite difference, while staggered Marker-And-Cell (MAC) for the Flow around obstacles, which is the same as the benchmark \cite{Wandel2020}. 
It is worth noting that the application of a higher-order finite difference method would incur additional computational costs during the calculation of the physics-constrained loss in the training process. However, it would not impact the inference stage of the model.

\subsection{Application in optimal control}
\label{app:freq_control}
\begin{figure}[!htb]
  \begin{center}
    \vspace{-0.3cm}
    \includegraphics[width=0.45\textwidth]{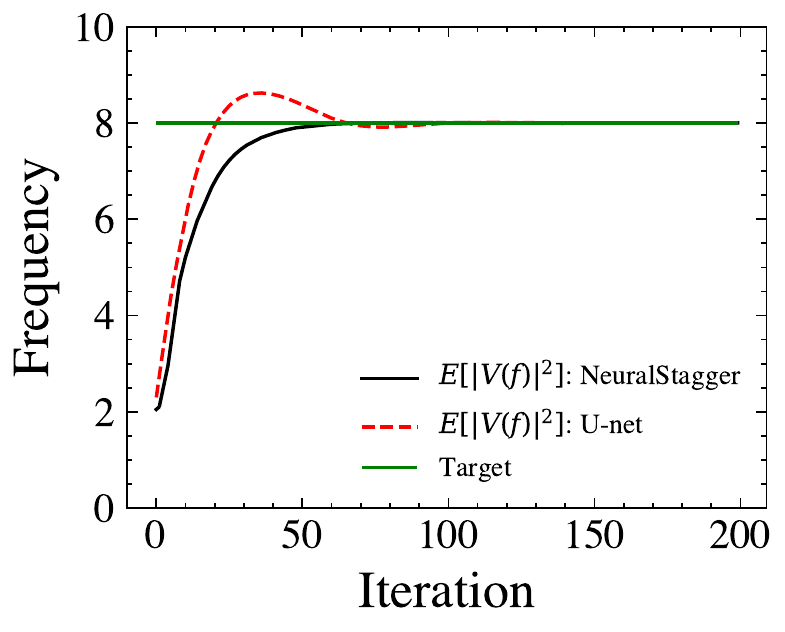}
  \end{center}
    \caption{The optimization curve of the frequency control for vortex streets. 
    The U-net converged after almost 72 iterations, while the 
    LordNet using NeuralStagger
    % with skipping factor $\{s_H:1,s_W:3,s_T:2\}$ 
    converged after 55 iterations.}
    \label{fig:frequency_control}
\end{figure}

The example of the inverse problem used in this paper is the same as the one in \citet{Wandel2020}. The goal is to change the flow speed to control the shedding frequency of a Kármán vortex street. 
The shedding frequency is estimated by the frequency spectrum
${V}(f)$ of the y-component of the velocity field behind the obstacle over 200 time steps, denoted by $E\left[\left|V(f)\right|^2\right]$.
We define the loss function $L=\left(E\left[\left|V(f)\right|^2\right] - \hat{f}\right)^2$, where $\hat{f}$ is the target frequency.
Then we compute the gradient of the velocity with regard to the loss by auto-differentiation through the neural solver and leverage Adam optimizer ~\citep{paszke2017automatic,kingma2014adam} to update the velocity.
We compare the result of the learned model with the setting $s_H=1, s_W=3, s_T=2$ to that shown in \citet{Wandel2020}.
As is shown in Figure~\ref{fig:frequency_control}, the velocity controlled by LordNet converges to the target velocity with fewer iterations. 

\subsection{The alleviation of the accuracy decrease from the spatial decomposition}
\label{app:feature_loss_imp}
Adding supplement information that could communicate the information between decomposed patches can alleviate the accuracy decrease due to the information loss. 
For example, vorticity, which describes the local spinning motion of the fluid, is defined by 
\begin{equation}
    \omega=\frac{\partial v_y}{\partial x}-\frac{\partial v_x}{\partial y}.
\end{equation}
It contains the communications between the velocity component in different directions, which is suitable to make up the lost information to improve the accuracy. Another way is to embed the coordinates using positional encoding \cite{vaswani2017attention} (PE) as the additional input feature and then add them to the input of the neural networks to improve the accuracy. 

Here we take the 2-dimensional flow around obstacles cases to test. For the LordNet, its capacity will decrease when applied to the coarse-resolution grids. For the FNO, the main feature is embedded into the Fourier domain, whose information is well-reserved if the physical fields are smooth and the samples are not aliasing. Both of them will introduce ambiguous factors to the demonstration. To focus on the demonstration of the help of supplement information for accuracy improvement, we select the Unet to test. As shown in Table~\ref{tab:flow_ap_2d_feature}, when there is no network capacity decrease, the accuracy decrease of the NeuralStagger can be improved by introducing supplement information. 
Notice that, for the way of PE, there is no increase in the parameter size. For the way using vorticity, the input channel will increase by 1, yielding the increase of the parameter and computational overhead, but they can be ignored because it just influences the first layer to fit the input size.

\subsection{The full results of three cases with different spatial-temporal factors}
\label{app:fulllist}
The full results of three cases with different spatial-temporal factors are presented in Tables~\ref{tab:nse_exp_fno2d}, \ref{tab:lid_exp_fno}, \ref{tab:fap_exp_all}, \ref{tab:3d-fao-full}.
\begin{table}[!htb]
    \centering
    \caption{Tests on Navier-Stokes equation with periodic boundary condition.}
    \setlength{\tabcolsep}{0.012\columnwidth}
    \begin{tabular}{c|c|c|c|c}
         \toprule
         Temporal & 
         \multicolumn{2}{c}{$L_T=2$,~~~(1,1)} &
         \multicolumn{2}{c}{$L_T=2$,~~~(2,2)} \\
         Skipping & Error-1 & Error-200 & Error-1 & Error-200. \\
        \midrule
         1 & 0.0000058 & 0.0011939 & 0.0000074 & 0.0016352 \\
         5 & 0.0000297 & 0.0012140 & 0.0000308 & 0.0016848 \\
         10 & 0.0000626 & 0.0013126 & 0.0000654 & 0.0016719 \\
         20 & 0.0001074 & 0.0011042 & 0.0001321 & 0.0017580 \\
         25 & 0.0001623 & 0.0013976 & 0.0001833 & 0.0017831 \\
         40 & 0.0002501 & 0.0012091 & 0.0002545 & 0.0012931\\
        \bottomrule
    \end{tabular}
    \label{tab:nse_exp_fno2d}
\end{table}

\begin{table}[!htb]
\centering
\caption{Tests on Navier-Stokes equation with Lid-driven cavity boundary condition.}
\setlength{\tabcolsep}{0.016\columnwidth}
\begin{tabular}{c|c|c|c|c}
\toprule
Temporal &
        \multicolumn{2}{c}{$L_T=27$,~~~(1,1)} &
         \multicolumn{2}{c}{$L_T=27$,~~~(2,2)} \\
         Skipping & Error-1 & Error-2700 & Error-1 & Error-2700. \\
         \midrule
         1 & 1.82 e-5 & 0.00228 &  1.78 e-5 & 0.00283\\
         27 & 3.76 e-4 & 0.00255 & 5.38 e-4 & 0.00390 \\
         54 & 8.78 e-4 & 0.00404 & 1.18 e-3 & 0.00420\\
         108 & 1.86 e-3  & 0.00461 & 2.30 e-3  & 0.00478 \\
\bottomrule
\end{tabular}
\label{tab:lid_exp_fno}
\end{table}
\begin{table}[!htb]
    \centering
    \caption{
The Performance of NeuralStagger with different decomposition factors and neural networks in the flow-around-obstacles setting. 
}
    \setlength{\tabcolsep}{0.011\columnwidth}
    \begin{tabular}{c|c|c|c|c|c}
         \toprule
        \multirow{2}{*}{Config} &
         Temporal & 
         Spatial &
         Folded & \multirow{2}{*}{Cylinder} & GMACs \\
         &
         factor & factors &pipe & &  per card\\
         \midrule
         \multirow{6}{*}{U-net} & - & - & 6.32 e-5 & 1.24 e-4 & 29.60\\
         &1 & (1, 3) & 1.94 e-4 & 2.84 e-4 & 9.76\\
         &1 & (2, 6) & 4.35 e-4 & 1.09 e-3  & 2.40\\
         &2 & (1, 3) & 6.76 e-4 & 9.12 e-4   & 4.88 \\
         &8 & (1, 3) & 6.43 e-4 & 2.02 e-3 & 1.22\\
         &16 & (1, 3) & 1.11 e-3 & 3.70 e-3 & 0.61\\
         \midrule
         \multirow{6}{*}{FNO} & - & - &4.01 e-4  & 4.54 e-4 & 18.51 \\
         &1 & (1, 3) &6.01 e-4  & 1.01 e-3 & 6.22\\
         &1 & (2, 6) & 1.81 e-3 & 3.27 e-3 & 1.69\\
         &2 & (1, 3) & 3.88 e-4 & 6.74 e-4 & 3.11\\
         &8 & (1, 3) & 5.23 e-4 & 2.19 e-3 & 0.78\\
         &16 & (1, 3) &  5.77 e-4& 3.49 e-3 & 0.39\\
        \midrule
         \multirow{7}{*}{LordNet}&
         - & - & 1.05 e-5 & 4.11 e-5 & 71.04\\
         &
         1 & (1, 3) & 2.21 e-5 & 8.97 e-5 & 19.84\\
         &
         1 & (2, 6) & 5.00 e-4 & 2.98 e-3 & 4.46 \\
         &
         2 & (1, 3) & 5.51 e-5 & 1.19 e-4 & 9.92 \\
         &
         8 & (1, 3)  & 3.93 e-4   & 6.60 e-4 & 2.48 \\
         &
         16 & (1, 3)& 3.47 e-4  & 8.55 e-4 & 1.24  \\
        \bottomrule
    \end{tabular}
    \label{tab:fap_exp_all}
\end{table}
\begin{table}[!htb]
    \centering
    \caption{The performance of NeuralStagger with different neural networks in the 3-dimensional flow-around-obstacles setting.}
    \setlength{\tabcolsep}{0.042\columnwidth}
    \begin{tabular}{c|c|c|c}
         \toprule
         \multirow{2}{*}{Network} & Spatial & PDE & GMACs  
         \\
         & factors & residual & per card \\
        \midrule
         \multirow{3}{*}{U-net3d} & - & 1.05 e-4 & 62.73\\
         & (2,2,2) & 2.85 e-4 & 7.84\\
         & (4,2,2)& 4.34 e-4 & 3.92\\
         \hline
         \multirow{3}{*}{FNO3d} & - & 1.15 e-4 & 21.89\\
         & (2,2,2)&2.17 e-4 & 2.80\\
         & (4,2,2)&4.07 e-4 & 1.42\\
         \hline
         \multirow{3}{*}{LordNet3d} & - & 1.01 e-4 & 73.86 \\
         & (2,2,2) & 4.54 e-4 & 8.23\\
         & (4,2,2) &4.67 e-4 & 3.99\\
        \bottomrule
    \end{tabular}
    \label{tab:3d-fao-full}
\end{table}

\subsection{Application to a larger-scale problem with large Reynolds number}
\label{app:large_re}
To showcase the effectiveness of the proposed method in handling larger-scale problems with high Reynolds numbers, we conducted experiments on the Navier-Stokes equation with periodic boundary conditions.
The resolution for this case was set to 256$\times$256, and the viscosity was set to 1e-4.
The results of these experiments are presented in Table~\ref{tab:nse_exp_fno2d_large}. 
In this particular case, which falls within the turbulence regime \cite{li_fourier_2020}, the results highlight the robust performance of NeuralStagger in turbulent flows. 
Regardless of the spatial and temporal decomposition used during training, the network's predictions achieved a similar level of accuracy.
\begin{table}[!htb]
    \centering
    \caption{Further tests on Navier-Stokes equation with periodic boundary condition.}
    \setlength{\tabcolsep}{0.005\columnwidth}
    \begin{tabular}{c|c|c|c|c|c|c}
         \toprule
         Temporal & 
         \multicolumn{2}{c}{$L_T=2$,~~~(1,1)} &
         \multicolumn{2}{c}{$L_T=2$,~~~(2,2)} &
         \multicolumn{2}{c}{$L_T=2$,~~~(4,4)}\\
         Skipping & Error-1 & Error-200 & Error-1 & Error-200 & Error-1 & Error-200. \\
        \midrule
         1 & 2.46 e-4	 & 5.01 e-2 & 1.93 e-4	&4.77 e-2&	2.38 e-4	&4.93 e-2 \\
         5 & 9.91 e-4&	4.67 e-2	&9.89 e-4	&4.66 e-2	&1.36 e-3	&5.00 e-2 \\
         10 & 2.87 e-3&	5.04 e-2&	2.67 e-3	&4.84 e-2&	3.03 e-3&	5.09 e-2	\\
        \bottomrule
    \end{tabular}
    \label{tab:nse_exp_fno2d_large}
\end{table}

To further demonstrate the effectiveness of the NeuralStagger rather than a trivial interpolation, we compared the predictions from NeuralStagger to the interpolation-based results. 
As for spatial decomposition, given the initial states and trained model with $s_T=1, s_H=s_W=2$, the prediction error is 1.93 e-4, while the error of the result via bilinear interpolation based on the coarse-resolution field is 1.29 e-2.
As for temporal decomposition, given a sequence of states $\hat{u}_{0,5}$ and the trained model with $s_T=5, s_H=s_W=1$, the average prediction error of predictions $\hat{u}_{6,4}$ is 9.88 e-4, while the average error of results via the bilinear interpolation based on the predicted states at 5 and 10 is 3.78e-3. 
These tests mean the extra information, which satisfies the equation, is contained in intermediate states.

\subsection{The limitations of scalability to the irregular meshes}
\label{app:lim}
As we mentioned earlier, one limitation of our work is that the decomposition is specifically defined on the regular meshes.
To address this limitation, there are several well-established numerical methods available to handle irregular domains, such as the finite volume method with a triangular mesh. 
To construct the physics-constrained loss in such cases, we could adopt a similar approach to constructing algebraic equations in those numerical methods. 
Specifically, the neural network takes the physical properties in each volume as input and predicts the properties at the next time step as output.
Regarding the decomposition algorithm, the key is to split the mesh into a fixed number of groups as evenly as possible, enabling the neural network to learn each subtask effectively. 
As it is unnecessary to strictly obey the graph coloring rules, i.e., no adjacent nodes have the same color, there is much flexibility in choosing or designing the coloring algorithm. 
Taking the classical greedy algorithm as an example, we can define a coloring order by depth-first search in the mesh and assign a legal color to a node each time. 
We can further reweigh the priority of color to the inverse of the assignment times of that color so that each group tends to have similar amounts of nodes.

\end{document}